Qinghao Zhang[a], Miao Ye[a], Hongbing Qiu[a], Yong Wang[b] and Xiaofang Deng[a]

[a]School of Information and Communication, Guilin University of Electronic Technology, Guilin, China; [b]School of Computer Science and Information Security, Guilin University of Electronic Technology, Guilin, China

Correspondence should be addressed to Miao Ye[a]; yemiao@guet.edu.cn

# A Novel Anomaly Detection Method for Multimodal WSN Data Flow via a Dynamic Graph Neural Network

Anomaly detection is widely used to distinguish system anomalies by analyzing the temporal and spatial features of wireless sensor network (WSN) data streams; it is one of critical technique that ensures the reliability of WSNs. Currently, graph neural networks (GNNs) have become popular state-of-the-art methods for conducting anomaly detection on WSN data streams. However, the existing anomaly detection methods based on GNNs do not consider the temporal and spatial features of WSN data streams simultaneously, such as multi-node, multi-modal and multi-time features, seriously impacting their effectiveness. In this paper, a novel anomaly detection model is proposed for multimodal WSN data flows, where three GNNs are used to separately extract the temporal features of WSN data flows, the correlation features between different modes and the spatial features between sensor node positions. Specifically, first, the temporal features and modal correlation features extracted from each sensor node are fused into one vector representation, which is further aggregated with the spatial features, i.e., the spatial position relationships of the nodes; finally, the current time-series data of WSN nodes are predicted, and abnormal states are identified according to the fusion features. The simulation results obtained on a public dataset show that the proposed approach is able to significantly improve upon the existing methods in terms of its robustness, and its F1 score reaches 0.90, which is 14.2% higher than that of the graph convolution network (GCN) with long short-term memory (LSTM).

Keywords: Wireless Sensor Networks; Dynamic Graph neural network; Graph attention network; GRU

# 1. Introduction

Wireless sensor networks (WSNs) are wireless ad hoc networks composed of large numbers of sensor nodes. A WSN can monitor and collect multimodal data (i.e., multiple physical measurement values) from the area in which the network is deployed in real time, and the network nodes can communicate with each other (Sun, 2015; Xue et al., 2021). Wireless sensors have the characteristics of convenient deployment and strong reliability. They have been widely used in many research fields, such as national defense security, natural environment monitoring (Xu & Liu, 2017), medical and health monitoring, and industrial and agricultural production monitoring (Wu et al., 2011; Zhu et al., 2011).

However, the unreliability of sensors due to faults may lead to the incorrect sensor records. When serious events such as fires or toxic gas leakage occurs, an abnormal sensor record has a serious impact. At this time, it is necessary to detect the anomaly state in the associated WSN in a timely manner and take emergency treatment measures to prevent deterioration due to the event. Anomaly detection plays a vital role in the safe operation of a WSN system. In recent years, with the rapid development of wireless communication and electronic technology, the numbers of sensor nodes in WSNs and the modal dimensions monitored by each node have expanded. Accordingly, the difficulty and performance requirements of anomaly detection are also increasing. Therefore, it is particularly essential to design an anomaly detection method that is suitable for the characteristics of WSN data and has better performance.

In the usual interpretation of anomalies in sample datasets, anomaly detection is generally posed as finding outliers that differ from most sample points (Jin et al., 2001). Many scholars have proposed different definitions of anomaly concepts. Ramaswamy et al. (2000) proposed an anomaly definition based on distance. Assuming that there are $m$ points in the given space, in which there are $n$ anomaly points and the number of neighbors considered by each point is *k*, *n* and *k* are hyperparameters that must be simultaneously set by users. The distance between each point and its *k* nearest neighbors is calculated as the evaluation distance. Then, all evaluation distances are sorted, and the points that correspond the top *n* ranked evaluation distances are regarded as abnormal points. In WSNs, abnormal nodes are single or multiple sample data points that deviate from the overall data distribution of the data samples collected by the sensor node due to the failure of this node in the system. Therefore, the abnormality of the sensor node can be detected by analyzing the abnormality of the sample dataset collected by this sensor node. An event node is defined as a node where an observed sensor value deviates from the overall data distribution due to a special event in the area (such as a sudden fire in the surrounding environment). A fault node (i.e., an error node) is defined as a node that cannot collect data normally due to its own fault or an external attack.

At present, WSN time-series data anomaly detection faces many complex difficulties, such as high dimensionality, multiple scales, long correlations and nonlinearity. The pivot method is the most commonly used approach for constructing the confidence intervals of unknown parameters in statistics. In (Peng et al., 2018), the confidence interval calculated from sensor data through the pivot method was combined with the current sensor reading for anomaly detection. However, this method fails to consider the location relationships between sensor nodes and does not make full use of the spatial dimension information of WSN nodes. The authors in (Fei & Li, 2015) used WSN data obtained at the current time to construct points in the given space. Then, all the points were divided into two clusters by k-means clustering, and anomaly detection was carried out by calculating the distances from the data points to the centroid. Similarly, this method does not consider the correlations between the spatial positions of the sensor nodes. The authors in (Jiang & Yang, 2018) studied the problem of anomaly detection in sensor networks with graph signal processing knowledge. First, a graph Fourier transform was carried out on a subgraph to extract the subgraph signal component. Then, the subgraph signal was used to determine the abnormal center node of the subgraph. However, this method examines one measurement value (single-mode), so it lacks the consideration of multiple measurement values (multimodal). The authors in (Lin et al., 2020) combined the knowledge of random matrix theory with anomaly detection, and the average spectral radius of the data matrix in a sliding window was calculated to determine whether an anomaly occurred by judging whether there was a sudden change in the average spectral radius. However, this method fails to decompose singular values when dealing with large data matrices.

In recent years, deep neural networks have made great achievements in the field of big data-driven approaches and have been widely used because of their excellent performance, which exceeds that of traditional nonmachine learning methods (Zhang et al., 2018). In (Li et al., 2020), an anomaly detection method was proposed by using a convolutional neural network (CNN) and a long-short-term memory network (LSTM) to predict network traffic data in a subsequent timestamp to estimate the probability of a system anomaly occurring at the subsequent time. The authors in (Li et al., 2019) studied the use of a generative adversarial network (GAN) for anomaly detection, and the authors in (Park et al., 2018) studied the use of variable autoencoding (VAE) in combination with LSTM for anomaly detection. However, these methods still do not make full use of the topology information between network nodes, so they do not use the features of the spatial dimension for anomaly detection.

How to apply the spatial features of the topology information possessed among the nodes of a WSN to anomaly detection is the key to achieving improved detection performance. Graph neural networks (GNNs) can extract "interesting" features from the adjacency matrices and attribute matrices of attribute networks, so they have the ability to reflect the spatial characteristics of WSN data. Common GNNs include graph convolution networks (GCNs) (Kipf & & Welling, 2016), graph attention networks (GATs) (Veličković et al., 2017), and the graph sampling aggregation network (GraphSage) (Hamilton et al., 2017). A GCN represents the input graph and convolutes it in the frequency domain through a Laplace matrix to realize the convolution operation on the topological graph. However, GCNs have the problem of oversmoothing, which makes anomalies less distinguishable and affects the performance of the associated algorithm. GraphSage can solve the problem by which a GCN needs to update the whole graph every iteration, and it can reduce the hardware overhead of the associated algorithm by sampling and aggregating nodes. However, the limitation of this method regarding the number of samples leads to the loss of the important local information of some nodes, and GraphSage also exhibits oversmoothing when the network contains too many layers. A GAT can solve the oversmoothing problem by aggregating neighbor nodes through an attention mechanism.

Due to the lack of labels of abnormal data in real-world datasets, most existing methods use unsupervised learning for anomaly detection. According to their optimization strategies, anomaly detection models are mainly divided into prediction and reconstruction approaches. In (Ding et al., 2019), a GCN was used to construct an encoder and a decoder to reconstruct an adjacency matrix and an attribute matrix, and the reconstruction error was used to estimate the anomaly score. The authors in (You et al., 2020) changed their GCN to a GAT on the basis of (Ding et al., 2019), thereby solving the problem of oversmoothing. The authors in (Deng & Hooi, 2021) modified their method on the basis of a GAT (Veličković et al., 2017) by adding a sensor representation vector to predict the sensor observation values at subsequent time stamps and calculate sensor anomaly scores, yielding improved accuracy. In (Zhao et al., 2020), an anomaly detection method combining the advantages of a prediction model and a reconstruction model was proposed. Two parallel GATs were used to extract spatial and temporal features and then divided into two branches to reconstruct the adjacency matrix and attribute matrix and predict the sensor readings at the subsequent time. The prediction error and reconstruction error were used to jointly estimate the anomaly score. In (Pei et al., 2021), a residual GCN was used to reconstruct the attribute matrix, and the residual matrix was used to calculate the anomaly score. Recently, an increasing number of studies have discussed anomaly detection via contrastive self-supervised training. There is no doubt that this is an excellent training method for anomaly detection tasks without ground-truth labels. The authors in (Audibert et al., 2020) designed a two-stage adversarial training-based anomaly detection model. In the first stage, two encoding models were committed to reconstructing the input data matrix. In the second stage, one encoding model was used as a generator to generate negative samples, and the other encoding model was used as a discriminator to identify the true and false samples. The authors in (Liu et al., 2021) studied an anomaly detection method combining Deep Graph Infomax (i.e. DGI) (Veličković et al., 2019) and GraphSage, their approach samples subgraphs, constructs positive and negative samples, and then uses a discriminator to conduct adversarial training with the positive and negative samples. The authors in (Guo et al., 2020) studied an anomaly detection framework called Dynamic-DGI, which uses the mutual information between corrosion subgraphs and the total input graph for adversarial training. However, WSN data in the real world are data streams containing three dimensions, node, mode, and time dimensions, and the anomaly detection methods described above do not consider the features of these three dimensions simultaneously. They only consider two dimensions, such as a matrix composed of multimodal readings obtained over a period of time or a matrix composed of the multinode readings of a certain measurement attribute (i.e., a single mode) over a period of time. These approaches do not comprehensively consider the correlations between times, modes and spatial node positions when conducting anomaly detection for a WSN data flow. In addition, the authors in (Qi et al., 2020) extracted time correlation features and spatial correlation features through a gated convolution network and a GCN, reconstructed the input KPI data and finally calculated the anomaly score. However, the GCN used in this method has the problem of oversmoothing, so the feature information is not fully extracted during the coding process, which limits the anomaly detection performance of the resulting model.

In fact, the data flow generated by a WSN can be regarded as multiple sensor nodes with a certain topology structure that collects the dynamic time-series data of multiple modes regarding the surrounding environment. Such a flow not only has the topology information of spatial locations but also the time-series temporal information of multiple modes. Therefore, a WSN data flow can be regarded as a model in a dynamic GNN. Notably, the current advanced methods based on GNNs do not fully consider the spatial features and the temporal features of WSN data flows and often ignore their multiple mode correlation features. A dynamic GNN is a critical model for fully expressing the spatial characteristics of a graph structure and the temporal characteristics of the multiple modes of each node.

To solve this problem, this paper proposes an anomaly detection method for WSN data flows based on a dynamic GNN. The form of the input data determines the difference between the data processing techniques of the traditional method and our method. The input of the traditional method is a two-dimensional matrix, which is composed of the readings of multiple modes over a period of time or the readings of multiple nodes regarding a certain measurement attribute (i.e., a single mode) in a period of time. The input of the method in this paper is a three-dimensional tensor that is composed of the readings of multiple modes of multiple sensor nodes over a period of time. The model designed in this paper comprehensively considers the three dimensions of a data flow: sensor nodes, modes and times.

The main innovations of this paper can be summarized as follows.

(1) Compared with the existing approaches, which only consider the multimodal multi-time data of a single node or the multi-node multi-time data with a single mode and lacks the comprehensive consideration of spatial and temporal factors regarding multiple sensor nodes, the anomaly detection method designed in this paper considers the temporal features of the multimodal multi-time data of a single node and historical data. At the same time, the fusion of temporal features and spatial features between multi-node locations is also considered.

(2) The proposed anomaly detection method for a WSN data flow based on a dynamic GNN is organized with temporal features on a single mode extraction module, a multimodal correlation feature extraction module, and spatial correlation features between the sensor node extraction modules. Graph attention mechanisms are used in all modules. Temporal feature extraction methods such as RNNs and LSTM have the problems of being unable to construct long-term dependencies and having slow training speeds due to their large numbers of parameters. The method in this paper solves the long-term dependence problem with a GRU. At the same time, it can predict all readings of all modes of all sensor nodes at the next timestamp and calculate the inference score at the current time through the deviation between the prediction value and the actual value to judge whether the associated node is abnormal.

(3) To verify the effectiveness of the anomaly detection method designed in this paper, we inject four types of anomalies into a real WSN dataset: zero-turn changes, sudden changes, slow numerical changes and fast numerical changes. We apply the designed anomaly detection method to this dataset to obtain its precision, recall and F1 score values. Extensive experimental results show that the F1 score of our method for WSN anomaly data is 90.3%, which is better than that of the common GCN-based anomaly detection method.

The rest of this paper is organized as follows. Section 2 presents the background and related work regarding attribute networks, GATs, and GRUs. Section 3 describes the proposed anomaly detection model for multimodal WSN data flows based on a dynamic graph attention network. We conduct extensive experiments to evaluate our method in Section 4. Section 5 concludes this paper.

## 2. Background and Related Work

### 2.1. Attribute Network

Suppose we are given an attribute network $G = (A,X)$, in which $A \in \mathrm{R}^{M \times M}$ represents the adjacency matrix of the network, $M$ represents the number of total nodes in the network, and $X \in \mathrm{R}^{M \times D}$ represents the attribute matrix of $M$ nodes. The attribute features of each node can be represented by a $D$-dimensional vector, and the $D$-dimensional vectors of $M$ nodes are combined to form the attribute matrix $X$ of the attribute network.

### 2.2. Graph Attention Network

A graph attention network (GAT) is an outstanding method for the graph embedding of attribute networks. As shown in Fig. 1, a GAT calculates the attention weight $\alpha_{ij}$ between the central node $h_i^{(l)}$ and its neighbor nodes $h_j^{(l)}$ according to their correlation based on an attention mechanism and adaptively assigns weights to different neighbors, greatly improving upon the expression ability of the graph neural network model and solving the oversmoothing problem of multilayer GCNs.

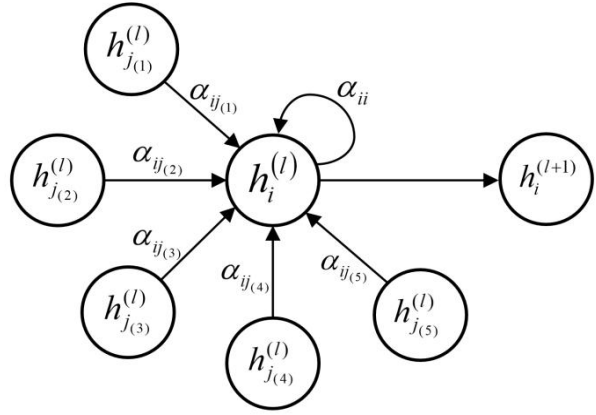

Figure 1. GAT calculation process.

Assuming that the given graph attention network has $L$ layers, the representation of node $i$ in the $l^{th}$ layer is $h_i^{(l)} \in \mathrm{R}^{F^{(l)}}$, where $F^{(l)}$ is the dimensionality of the representation of node $i$ in the $l^{th}$ layer. Assuming that the total number of nodes is $N$, $H^{(l)} = \{h_1^{(l)}, h_2^{(l)}, ..., h_N^{(l)}\}$ is the input of the $l^{th}$ layer, and the output of the $l^{th}$ layer is $H^{(l+1)} = \{h_1^{(l+1)}, h_2^{(l+1)}, ..., h_N^{(l+1)}\}$. The representation of node $i$ in the $l+1^{th}$ layer is $h_i^{(l+1)} \in \mathrm{R}^{F^{(l+1)}}$, where $F^{(l+1)}$ is the dimensionality of the representation of node $i$ in the $l+1^{th}$ layer. Note that the representation at the $0^{th}$ layer $H^{(0)} = X = \{x_1, x_2, ..., x_N\}$ is the attribute matrix of the attribute network.

First, we use a weight matrix $B \in \mathrm{R}^{F^{(l+1)} \times F^{(l)}}$ to map the representation vector of the input node $i$ in the $l^{th}$ layer to the hidden layer vector $q_i^{(l)}$.

$$q_i^{(l)} = Bh_i^{(l)} \tag{1}$$

Second, the correlation coefficient between node $i$ in the $l^{th}$ layer and node $j$ in the $l^{th}$ layer can be computed as in (2).

$$e_{ij}^{(l)} = \text{Leaky ReLU}\left(a^{(l)T}\left[q_i^{(l)} \| q_j^{(l)}\right]\right) \quad (2)$$

where $a^{(l)} \in R^{2F^{(l+1)}}$ is also the weight vector in the $l^{th}$ layer and node $j$ is the neighbor node of node $i$. From the adjacency matrix $A$, we can easily obtain the full neighbor node information of a certain node. $\|$ represents the combination operation. $e_{ij}^{(l)}$ indicates the correlation coefficient between node $i$ and node $j$ in the $l^{th}$ layer. Then, the correlation coefficient calculated between each node and all related neighbors is normalized by a *softmax* function to obtain the attention weight matrix.

$$\alpha_{ij}^{(l)} = \frac{\exp\left(e_{ij}^{(l)}\right)}{\sum_{v_k \in N(v_i)} \exp\left(e_{ik}^{(l)}\right)} \quad (3)$$

where $\alpha_{ij}^{(l)}$ represents the attention weight coefficient between node $i$ and node $j$ in the $l^{th}$ layer. $N(v_i)$ represents the set of neighbor nodes of node $i$. The normalization process in (3) ensures that the sum of the attention weight coefficients between every node and all its neighbor nodes is 1.

Finally, the representation $h_i^{(l)} \in R^{F^{(l+1)}}$ of node $i$ in the $l+1^{th}$ layer is calculated by using attention weight coefficients and a hidden layer vector, where $\sigma$ is a nonlinear activation function.

$$h_i^{(l+1)} = \sigma\left(\sum_{j \in N_i} \alpha_{ij}^{(l)} q_j^{(l)}\right) \quad (4)$$

The dimensions of $q_i^{(l)}$ and $q_j^{(l)}$ are $F^{(l+1)}$, while $\alpha_{ij}^{(l)}$ is a constant. The dimensionality of the representation of node $i$ in the $l^{th}$ layer is $F^{(l)}$, and the dimensionality of the representation of node $i$ in the $l+1^{th}$ layer is $F^{(l+1)}$. In other words, the GAT transforms the $d$-dimensional features of the current node into another $d'$-dimensional feature representation by aggregating the information of neighbor nodes.

*2.3. GRU*

Recurrent neural networks are often used to extract the temporal features of time-series data. However, traditional RNNs have difficulty modeling long-term dependencies. An effective scheme is to introduce a gate structure based on an RNN to control the propagation of information.

The long-term and short-term memory network (LSTM) controls the propagation of information by introducing three gates, namely, a forget gate, an input gate and an output gate. The structure of the gated recurrent unit (GRU) (Chung et al., 2014) is simpler than that of LSTM. A GRU has only two gates: an update gate and a reset gate. The calculation formulas are as follows.

$$z_t = \sigma(W_z x_t + U_z h_{t-1}) \quad (5)$$

$$r_t = \sigma(W_r x_t + U_r h_{t-1}) \quad (6)$$

$$\tilde{h}_t = \tanh(W_h x_t + U_h (r_t \circ h_{t-1})) \quad (7)$$

$$h_t = z_t \circ h_{t-1} + (1 - z_t) \circ \tilde{h}_t \quad (8)$$

The input of a GRU consists of two parts: the input tensor $x_t$ of the current time $t$ and the hidden layer state $h_{t-1}$ of the previous time $t-1$. The overall framework is illustrated in Fig. 2. All the symbols in $\{W_z, W_r, W_h, U_z, U_r, U_h\}$ form the weight matrix. Through the calculation of the above formula, the state of the hidden layer at the current time is finally obtained. A GRU has fewer parameters than LSTM, so it converges more easily and is more suitable for building larger networks.

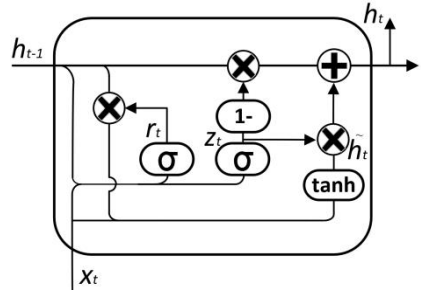

Figure 2. GRU framework.

## 3. Anomaly Detection Model for a Multimodal WSN Data Flow Based on a Dynamic Graph Attention Network While Considering Spatiotemporal Features

### 3.1. Multimodal WSN Data Flow Model Considering Spatiotemporal Features

When discussing the correlation features between different nodes and different modes of a data flow, a WSN data flow model should be established first.

Assume that a wireless sensor network is deployed in a certain target area that contains $M$ sensor nodes, and $N$ sensors are deployed in each sensor node to monitor the region's temperature, humidity, carbon dioxide concentration and other modal indicators. A time synchronization mechanism can be used to ensure the synchronization of information transmission and data collection between the nodes in the WSN.

To make full use of the historical data and current observation data for the real-time analysis of data features, we adopt a sliding window model to build a data flow model (Datar et al., 2002). Assuming that the current timestamp is $t$ and that the length of the window is set to $W$, the data flow sample points of all sensors in all sensor nodes within the time step $\{t-W-1, t-W,...,t-1\}$ are truncated to form a three-dimensional data flow tensor $X \in R^{M \times N \times W}$. The data of the leftmost element expires when a new data element enters from the right, and the window slides the length of one data element to the right. For example, the time range of the data flow model at time $t$ is $\{t-W-1, t-W,...,t-1\}$, and the time range of the data flow model at time $t+1$ is $\{t-W, t-W+1,...,t\}$. Therefore, the WSN data flow can be regarded as a dynamic network that changes with time. Fig. 3 shows the visual model of a three-dimensional data flow, in which each node can collect time-series data regarding multiple attributes (modes), such as temperature (temp), carbon dioxide (CO2) content, and oxygen (O2) content.

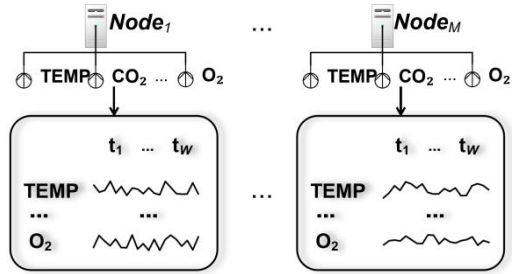

Figure 3. Three-dimensional data flow.

### 3.2. Overview of the Abnormal WSN Node Detection Model

The structure of the anomaly detection model is shown in Fig. 4. It is mainly composed of five modules: a preprocessing module, a multimodal data correlation feature extraction module for single sensor node, a temporal data feature extraction module for single sensor node, a dimensionality reduction module for the multimodal temporal data of single sensor node, and a spatial correlation feature extraction module between the positions of sensor nodes. Here, $m_i$ is the $i^{th}$ node and $m_j$ is a neighbor node with its spatial correlation; $n_i$ is the $i^{th}$ mode, and $n_j$ is a neighbor mode with its correlation; $t_i$ is the $i^{th}$ timestamp, and $t_j$ is a neighbor timestamp with its temporal correlation.

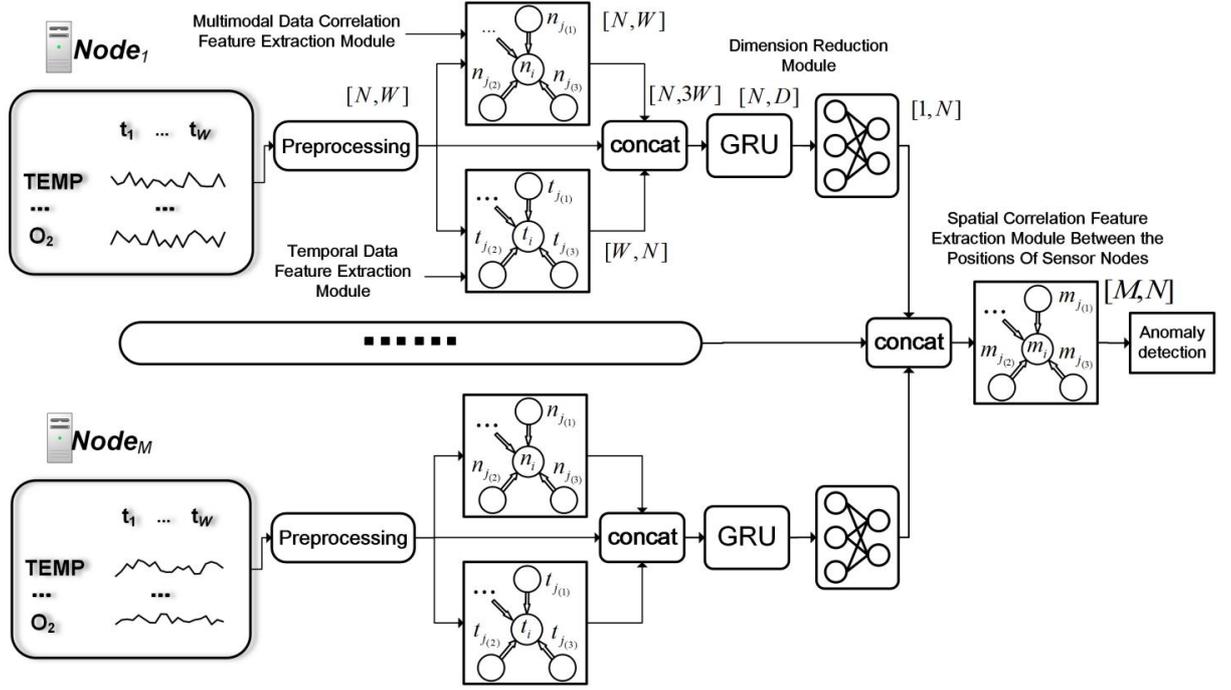

Figure 4. Anomaly detection framework.

  In the anomaly detection model, each sensor node leads out along a branch to standardize the data preprocessing approach for the multimodal and time-series data flow of each sensor node. Then, two GATs are used to extract the correlation features of different modes and the temporal features at different timestamps from the preprocessed data flow. Next, the outputs of the two GATs are combined with the preprocessed data as the input of the dimensionality reduction module, and the feature representation vector of the current sensor node is obtained after dimensionality reduction processing is completed by the GRU. When all branches of all sensor nodes have been processed, the representation vectors of all sensor nodes are combined as the input of the node spatial feature extraction module. Then, the complete graph representation tensor is obtained by the node spatial feature extraction module through a group of GATs. Finally, the graph representation tensor is used to predict all readings of the WSN at the current time.

  The specific design of each module and its functions are described in detail below.

## 3.3. Preprocessing Module

Different measurement attributes (modes) often have different measurement ranges, so the orders of magnitude between different modes vary greatly. If the original data are directly used for analysis, the roles of modes with higher values in the comprehensive analysis will be strengthened, and the role of modes with lower values in the comprehensive analysis will be relatively weakened. Therefore, to eliminate the influence of the measurement range differences between modes, it is necessary to standardize the original data so that each mode lies in the same value range. In this paper, the Z score standardization method is used to standardize the input tensor. The formula is as follows:

$$\tilde{X}_{ij} = \frac{X_{ij} - \mu(X_{ij})}{\sigma(X_{ij})} \tag{9}$$

where $X \in R^{M \times N \times W}$ is the input tensor, $X_{ij}$ represents the time-series data measured by the $j^{th}$ sensor of the $i^{th}$ node, $\mu(X_{ij})$ represents the mean value of $X_{ij}$ and $\sigma(X_{ij})$ represents the standard deviation of $X_{ij}$. After Z score preprocessing, the multimodal multi-time data conform to a normal distribution with a mean value of 0 and a standard deviation of 1.

## 3.4. Multimodal Data Correlation Feature Extraction Module

This module uses a graph attention network to extract the correlation features between the different modes of a multimodal multi-time data flow.

Before building the dependence relationships between the modes, the correlation coefficients between data vectors should be introduced first, and the time-series data of a mode is also a kind of data vector. Assuming that $P$ and $Q$ are two known $n$- dimensional data vectors, the correlation coefficients of these two data vectors can be computed by formula (10). Each correlation coefficient is between -1 and 1.

$$\rho_{PQ} = \frac{Cov(P,Q)}{\sqrt{Var[P]\,Var[Q]}}$$

$$= \frac{\sum_{i=1}^{n}(P_i - \bar{P})(Q_i - \bar{Q})}{\sqrt{\sum_{i=1}^{n}(P_i - \bar{P})^2 \cdot \sum_{i=1}^{n}(Q_i - \bar{Q})^2}} \quad (10)$$

When we possess prior information (i.e., when we know whether there are some dependencies between the modes, this information can be flexibly represented as a set of dependency relationships. Suppose that $C_i$ is used to represent the set of all other modes on which the $i^{\text{th}}$ mode depends. The adjacency matrix $A \in \mathrm{R}^{N \times N}$ needed for the input of the GAT in this section can be constructed by $C_i$, where $V_i$ and $V_j$ are the time-series data vectors of mode $i$ and mode $j$, respectively.

$$A_{ji} = \begin{cases} \rho_{V_i V_j} & j \in C_i \\ 0 & j \notin C_i \end{cases} \quad (11)$$

When there is no prior information (i.e., when we do not know the dependencies between the modes), the adjacency matrix $A \in \mathrm{R}^{N \times N}$ required in the input of the GAT in this section can be set as the full-1 matrix, which means any mode regards all other modes (including itself) as neighbor modes. Based on the attention mechanism of the GAT, the attention weight coefficients between the modes can be adaptively learned, and the information of the neighbor modes can be aggregated according to their attention weight coefficients.

The attribute matrix required by the GAT in this section is $\tilde{X}_i \in R^{N \times W}$. The GAT on the branch of node $i$ in this section inputs the preprocessed multimodal multi-time matrix as the attribute matrix, where $N$ represents the number of modes and $W$ represents the length of the sliding window. As mentioned in Section 2.2, from the input layer to the output layer, the GAT aggregates the information of the attribute matrix according to the structural matrix (adjacency matrix) and transforms the dimensionality of the attribute matrix. The dimensions of the input layer and output layer of the GAT in this module are set to the same value, so the output of the GAT in this module is a matrix of shape $N \times W$.

### 3.5. Temporal Data Feature Extraction Module

This module uses a graph attention network to extract the correlation features between the timestamps of a multimodal time-series data flow.

$$U_{ij} = 1 \quad \{U \in R^{w \times w}\} \tag{12}$$

$$A = U - E \tag{13}$$

The adjacency matrix $A \in R^{W \times W}$ of the GAT in this section can be computed by formulas (12-13), where $U$ represents the full-1 matrix and $E$ represents the identity matrix. Any moment regards all other moments as neighbors. Based on the attention mechanism of the GAT, the attention weight coefficients between the timestamps can be adaptively learned, and the information of neighbor timestamps can be aggregated according to their attention weight coefficients.

The attribute matrix required for the attention network is $\tilde{X}_i^T \in R^{W \times N}$. The GAT on the branch of node $i$ in this section inputs the transpose of the preprocessed multimodal multi-time matrix as the attribute matrix, where $N$ represents the number of modes and $W$ represents the length of the sliding window. As mentioned in Section 2.2, from the input layer to the output layer, the GAT aggregates the information of the attribute matrix according to the structural matrix (adjacency matrix) and transforms the dimensionality of the attribute matrix. The dimensions of the input layer and output layer of the GAT in this module are set to the same value, so the output of the GAT network in this module is a matrix of shape $W \times N$.

### 3.6. Dimensionality Reduction Module for the Multimodal Temporal Data of a Single Sensor Node

In Sections 3.4 and 3.5, we describe two of GATs that are used to learn the complex dependencies of multi-node multi-time data in both the temporal and modal dimensions. The shape of the final output matrix of the GAT network that focuses on extracting intermodal features is $N \times W$, and the shape of the final output matrix of the GAT network that focuses on extracting intertemporal features is $W \times N$. The output matrices of the two GATs and the preprocessed data matrix are spliced together to form a characteristic matrix with a shape of $N \times 3W$. Then, the dimensionality of the characteristic matrix can be reduced to $N \times D$ through the GRU module introduced in Section 2.3. Finally, the feature matrix is transformed into a sensor node representation vector with $N$ dimensions through the fully connected layer.

### 3.7. Spatial Correlation Feature Extraction Module Between the Positions of Sensor Nodes

In this module, the spatial correlation features of multiple nodes are extracted from the node dimension and modal dimension through the graph attention network.

When $M$ sensor nodes are contained in the network, $M$ branches correspond to these nodes in the model, and the representation vector of the corresponding sensor node is obtained at the end of each branch. The attribute matrix input by the GAT network in this section is the $M \times N$ matrix composed of the representation vectors of all sensor nodes. The input structural matrix $A$ is the adjacency matrix of the $M$ nodes in the network. Similarly, the dimensions of the input layer and output layer of the GAT network in this module are set to the same value, so the final output of the model is a matrix with a shape of $M \times N$.

We calculate the root mean squared error (RMSE) of the final output matrix and the current observed readings for $N$ modes of $M$ nodes and use it as the loss function.

$$L_{MSE} = \frac{1}{T_{train} - w} \sum_{t=w+1}^{T_{train}} \left( \tilde{s}(t) - s(t) \right)^2 \qquad (14)$$

where $\tilde{s}(t)$ is the predicted matrix output of the model at time t and $s(t)$ is the reading matrix for the $N$ modes of $M$ nodes observed at time $t$. $w$ represents the length of the sliding window, and $T_{train}$ is the total duration of the training set.

### 3.8. Inference Score Calculation

We calculate the node anomaly score of each sensor node in the test set and filter out the largest value as the inference score at the current time $t$. We can judge anomalies at the current time and locate the anomaly nodes through their inference scores.

$$score_i(t) = \sum_{j=1}^{N} \left( s(t)_{ij} - \tilde{s}(t)_{ij} \right)^2 \qquad (15)$$

$$score(t) = \max_i score_i(t) \qquad (16)$$

where $score_i(t)$ represents the node anomaly score of node $i$ at time $t$. $s(t)_{ij}$ represents the observation value of mode $j$ of node $i$ at time $t$, and $\tilde{s(t)}_{ij}$ represents the model output prediction value of mode $j$ of node $i$ at time $t$. $score(t)$ represents the inference score.

Finally, if $score(t)$ exceeds the preset threshold, the status of the wireless sensor network at time $t$ is regarded as an anomaly. To avoid introducing extra hyperparameters, we use a simple method to determine the value of the threshold, import the validation set data into the model to obtain the inference scores of the validation set, and finally set the threshold to the maximum inference score of the validation set.

## 4. Experiments

### *4.1. Experimental Setup*

To prove the feasibility of the anomaly detection method proposed in this paper, this section carries out model training, anomaly detection and performance verification on WSN sensor data through experiments. The hardware environment used for the experiment is a server with an Intel Xeon silver 4210 CPU @2.20 GHz and an NVIDIA GeForce RTX 2080Ti GPU. The software environment used for the experiment is Python. We implement our method and its variants in the pytorch-1.9.0 deep learning framework and the Matplotlib drawing library. The version number of CUDA is 11.1.

### *4.2. Dataset and Abnormal Point Injection*

In this paper, we use the wireless sensor network dataset deployed by the Intel Berkeley Research Lab (Madden, 2004) to verify the effectiveness of our model. The IBRL sensor network is composed of 54 sensor nodes, each of which collects four modal attributes: temperature, humidity, light intensity and voltage. The data collection period ranged from February 28, 2004, to May 5, 2004. The spatial locations of the sensor nodes are shown in Figure 5.

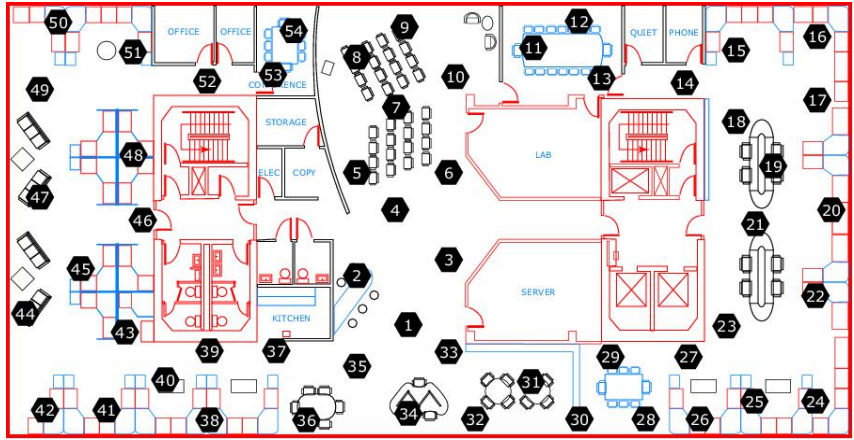

Figure 5. Spatial locations of sensor nodes.

In the experiment, we select three modes of data (temperature, humidity and voltage data) from 50 sensor nodes at 3000 moments from March 4 to March 7 to form the original data tensor with a shape of $50 \times 3 \times 3000$. Because the amount of abnormal data is very small in the actual system, this paper generates labeled abnormal data samples by manually injecting various types of abnormal points.

Abnormal data samples are obtained through the injected anomaly calculation functions in Table 1 to cover the original normal data samples. The generation modes of the four types of anomalies in the injected anomaly calculation function are described as follows: $\tau$ is the duration of the anomaly, $X_{ij}(t)$ is the reading of the $j^{th}$ mode of sensor node $i$ at time $t$, $X_j^{max}$ is the maximum value of the $j^{th}$ mode measured in the training set, and $X_j^{min}$ is the minimum value of the $j^{th}$ mode measured in the training set.

Anomaly type 1 simulates the slow mode changes in the environment detected by a node that deviates from its conventional state (e.g., the observed value of the gas content of a sensor node gradually increases due to the leakage of toxic gas in the chemical plant).

Anomaly type 2 simulates the fast mode changes in the environment detected by a node (i.e., the temperature observation of a sensor node rises rapidly in a short period of time due to a fire).

Anomaly type 3 simulates the extreme changes in the external environment detected by a node (i.e., the sudden invasion of high-temperature sources leads to sudden changes in the temperature observations to a certain extent).

Anomaly type 4 simulates the failure of a sensor node due to external interference, and the observed value of the mode becomes 0.

Table 1. Injected Anomaly Calculation Functions.

| Anomaly type | Function | During |
|---|---|---|
| 1 | $X_{ij}(t) = X_{ij}(t) \pm (X_j^{max} - X_j^{min})/p$  ($p = 14$) | $[1,\tau]$ |
| 2 | $X_{ij}(t) = X_{ij}(t) \pm (X_j^{max} - X_j^{min})/q$  ($q = 9$) | $[1,\tau]$ |
| 3 | $X_{ij}(t) = X_{ij}(t) \pm (X_j^{max} - X_j^{min})$ | $[1,\tau]$ |
| 4 | $X_{ij}(t) = 0$ | $[1,\tau]$ |

In this experiment, the ratio of the training set, verification set, and test set is 8:1:1. The training set contains 2400 sampling moment data for three modes of 50 sensor nodes. The verification set and the test set each contain 300 sampling moment data for the three modes of 50 sensor nodes.

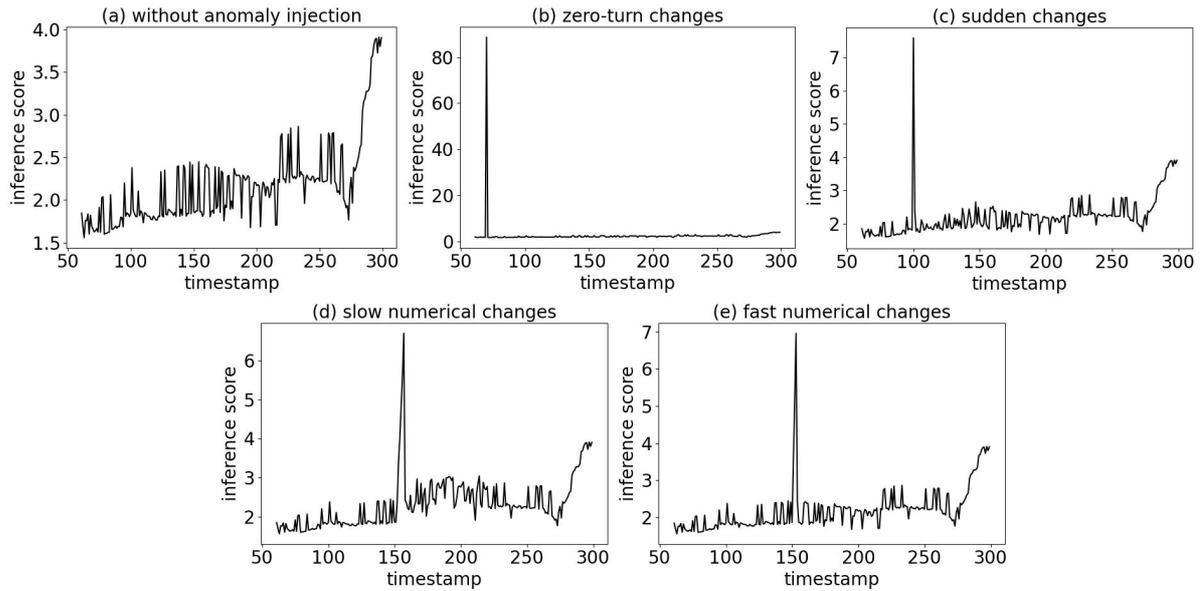

Figure 6. Changes in the inference scores inferred by the model when four types of anomalies are injected.

Fig. 6 depicts the changes in the inference scores inferred by the model when four types of anomalies are injected. Since the length of the sliding window is 60, the inference score is calculated from the $61^{st}$ timestamp. Subgraph (a) is the inference score curve for the test set without anomaly injection. Subgraph (b) is the inference score curve for the test set injected with a type 4 abnormality at the $70^{th}$ sampling moment of the voltage mode of node 29. Subgraph (c) is the inference score curve for the test set injected with a type 3 abnormality at the $100^{th}$ sampling moment of the temperature mode of node 30. Subgraph (d) is the inference score curve for the test set injected with type 1 abnormalities at the $150^{th}$-$157^{th}$ sampling moments of the humidity mode of node 35. Subgraph (e) is the inference score curve for the test set injected with type 2 abnormalities at the $150^{th}$-$153^{rd}$ sampling moments of the temperature mode of node 23. After the injection of anomalies, the numerical distribution deviates from the original distribution, which expands the difference between the modal observation value and the model prediction value and then enlarges the inference score.

*4.3. Evaluation Metrics*

We use precision (Prec), recall (Rec) and F1 score (F1) metrics calculated over the test set to evaluate the performance of our method, and TP, TN, FP and FN represent the four values of the confusion matrix. $F1 = \frac{2 \times Prec \times Rec}{Prec + Rec}$, $Prec = \frac{TP}{TP + FP}$ and $Rec = \frac{TP}{TP + FN}$.

Assuming that the sampling moment set with a length of *T* is extracted from the test set, 50% of the sampling moments set are extracted as Set-1, and the remaining 50% of the moments are extracted as Set-2. The moments in Set-1 are the timestamps that need injected anomalies for testing, and the moments in Set-2 are the timestamps that do not need injected anomalies for testing.

For the whole test set cycle, *T* anomaly detection iterations are carried out, with one type of abnormality or no abnormality injected into the test set every time according to the timestamps in Set-1 or Set-2.

Assuming that the sampling time of the injected anomaly is *t* and that the allowable delay constant is *delaystep*, if the inference score exceeds the threshold within the time range [*t, t + delaystep*], it is regarded as a detected injected anomaly (TP); if the inference score fails to exceed the threshold within the time range [*t, t + delaystep*], it is considered that there is an abnormality that has not been detected (FP); however, if the inference score exceeds the threshold when there is no abnormality actually within the time range [*t, t + delaystep*], it is regarded as no abnormality but detected as an abnormality (FN). The value of *delaystep* is set to 8 in this paper.

## 4.4. Sensitivity Test

We test the sensitivity of our anomaly detection method to anomalous data (the impact of injected anomaly data with different deviations regarding the precision of anomaly detection).

The deviation extent of the injected anomaly and the threshold affect the resulting F1 score. The greater the deviation of the injected abnormal data from the original data, the greater the inference score is; thus, it is easier to exceed the threshold, and the precision increases accordingly. When the threshold is lower, an inference score more easily exceeds the threshold, and the precision increases, but it also becomes easier to cause misjudgments and reduce the recall of the model. As mentioned in Section 3.8, the threshold setting method in this paper is to set the maximum inference score obtained on the verification set as the threshold. Next, we study the relationship between the deviation extent of the injected anomaly and the precision of anomaly detection.

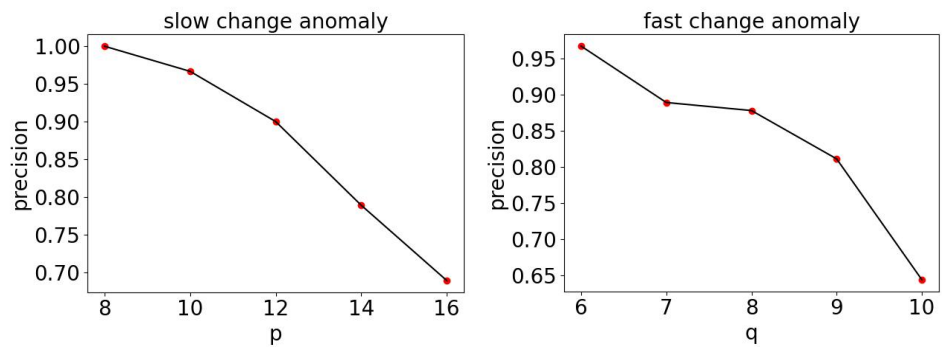

Figure 7. Deviation extents of injected anomalies and the precision of anomaly detection.

Since anomaly type 3 and anomaly type 4 do not contain hyperparameters, Fig. 7 only shows the detection precision curves obtained for slow change anomalies and fast change anomalies with different injection deviations (i.e., the $p$ value and $q$ value in Table 1 must be changed to change the deviation of the injected anomaly). We find that when the values of $p$ and $q$ gradually increase, the deviation extent of the injected anomaly decreases, and the inference score more difficult to exceeds the threshold, which results in a decline in precision. In subsequent experiments, the slow change injected anomaly parameter $p$ is set to 14, the duration $\tau$ is 8, the fast change injected anomaly parameter $q$ is set to 9, the duration $\tau$ is 4, the duration $\tau$ for the sudden change and zero-turn anomalies is 1, the total number of training epochs is 60, the model optimizer is set to Adam, and the learning rate is set to 0.00005.

Through the sensitivity test experiment, we find that the voltage mode is the most sensitive to small data changes among the three modes, and a small change in the original real voltage data may cause a surge in the inference scores. Among the four injected anomaly modes, the zero-turn anomaly is the most sensitive. The zero-turn anomaly often causes a significant surge in the inference scores (far exceeding the threshold), followed by the sudden anomaly. The sensitivity of fast change anomalies and slow change anomalies is weaker than that of sudden anomalies. This section determines the hyperparameters in the injected anomaly calculation function. The next section is carried out to determine the other parameters in the anomaly detection model.

## 4.5. The Influence of the Sliding Window Length and GRU Hidden Layer Parameters on the Performance of the Anomaly Detection Method

Sliding windows with different lengths have different effects on the statistical characteristics of data flows. The size of the window also directly affects the anomaly detection speed of the resulting model because the detection speed is related to the duration of the window. As shown in the sliding window test in Fig. 8, we conduct experiments on the test set with sliding window lengths K ∈ [30, 40, 50, 60, 70] and obtain metric values under these different lengths.

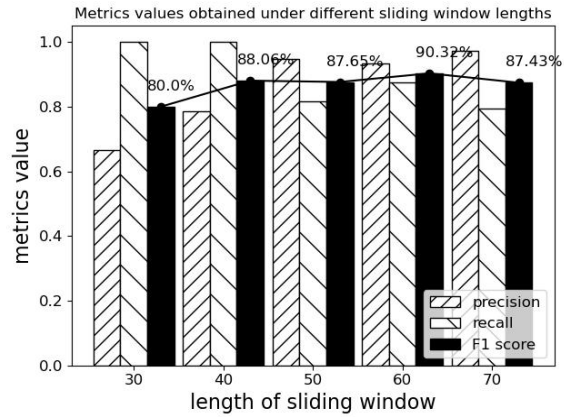

Figure 8. Metrics values obtained under different sliding window lengths.

With increasing window length, the precision increases, and the corresponding recall decreases. Based on the analysis of the results in terms of three performance indicators, to maintain the relative balance between precision and recall, it is appropriate to set the length of the sliding window to 60. In subsequent experiments, the sliding window length is set to 60.

The use of too few neurons in the GRU hidden layer leads to underfitting. In contrast, the use of too many neurons in the hidden layer increases the training time and causes overfitting. In addition, each node branch in this model involves the GRU module, so increasing the number of neurons in the hidden layer increases the inference time required by the model and decreases the speed of anomaly detection. Notably, it is essential to select an appropriate number of hidden layer neurons. The number of GRU layers in this paper is set to 2. Fig. 9 depicts the metric values obtained under different numbers of GRU hidden layer neurons.

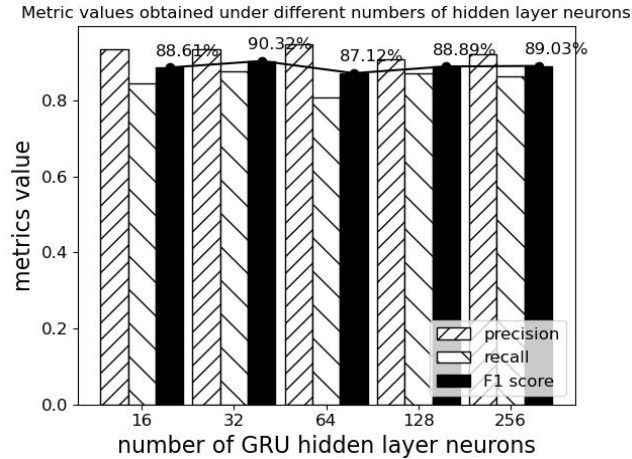

Figure 9. Metric values obtained under different numbers of GRU hidden layer neurons.

The performance of the model improves with the increase in the number of hidden layer neurons. However, the performance of the model begins to decline after the number of neurons reaches a certain value. The reason for this finding is that the gated recurrent unit cannot be trained completely in a limited number of epochs. Therefore, in subsequent experiments, we set the number of hidden layer neurons to 32.

*4.6. Comparison with Existing Methods*

Due to the lack of research on anomaly detection models that are suitable for multi-node, multimodal and multitemporal WSN data flows, this paper conducts a comparative experiment by changing the internal module of the model into other common modules.

Experimental scenario 1: We replace the graph attention network in the model with a graph convolution network (GCN). At the same time, we also replace the gated recurrent unit in the model with a long short-term memory network (LSTM).

Experimental scenario 2: We modify the Z score normalization method used in the model to the maximum and minimum normalization methods. The upper and lower limits of the three modes can be obtained from the dataset. The values of temperature and humidity range from 0-100, and the voltage values range from 2-3.

Experimental scenario 3: Two methods can be used to construct the adjacency matrix of the node dimension. One solution is to use the full-1 matrix as an adjacency matrix, which means that the neighbor nodes of any sensor node are composed of all other sensor nodes (including self-connections). The second solution is to use the TOPK nearest neighbor method to construct an adjacency matrix. The spatial location of the wireless sensor network is shown in Fig. 5. The dataset contains the *X* and *Y* coordinate information of all sensor nodes in the two-dimensional plane space.

We construct the TOPK adjacency matrix via formulas (17-18). $distance_{ij}$ represents the distance from node $i$ to node $j$, and $x_i$ and $y_i$ represent the *x* and *y* coordinates of node $i$ in the two-dimensional space, respectively. $distanceTopK^i_{min}$ represents the set of *k* nodes that are closest to node $i$. Each node only constructs edge connections with the nearest *k* nodes in its space and does not form edge connections with other nodes.

$$distance_{ij} = \sqrt{(x_i - x_j)^2 + (y_i - y_j)^2} \qquad (17)$$

$$A_{ij} = \begin{cases} 1 & j \in distanceTopK^i_{min} \\ 0 & j \notin distanceTopK^i_{min} \end{cases} \qquad (18)$$

The experimental results are shown in Table 2. It can be found that the max-min normalization used in experimental scenario 2 achieves the maximum precision on the test set, while its recall is the minimum value among those of all methods. The reason for this is that the maximum inference score obtained on the verification set is close to the minimum inference score obtained on the test set. As mentioned in Section 3.8, the anomaly detection threshold of this method is set as the maximum inference score value of the verification set. When the inference score of the test set exceeds the threshold at a certain time, the system status is regarded as anomalous. When the threshold is close to the minimum inference score value obtained on the test set, most of the moments in the test set are directly judged as anomalies. As a result, the precision surges, and the normal moments also tend to be judged as anomalies, resulting in a lower recall. When using the full-1 matrix to construct the adjacency matrix in this paper, the precision and recall are relatively balanced. When using the TOPK method to construct the adjacency matrix, the precision is the maximum value among those of all methods. The comprehensive performance achieved by constructing the adjacency matrix with the full-1 matrix is the best among all methods, and the F1 score is increased by 4.9% over that of the TOPK adjacency matrix construction method. Compared with those of the GCN-LSTM method, the precision and recall of our model are improved by 8% and 18.7%, respectively, and the F1 score increases by 14.2%.

Table 2. Comparison of Experimental Results.

| method | Prec | Rec | F1 |
|---|---|---|---|
| GCN-LSTM | 85.3% | 68.8% | 76.1% |
| MAXMIN | **100%** | 51.4% | 67.9% |
| GAT-GRU(Full-1) | 93.3% | **87.5%** | **90.3%** |
| GAT-GRU(TopK) | 98.8% | 75.2% | 85.4% |

### 4.7. Effectiveness of Graph Attention Networks

To study the effectiveness of each graph attention network component in the developed model, this paper uses the exclusion method to individually disable the modules in the model that involve GATs to observe how the anomaly detection performance changes.

First, we evaluate the effectiveness of multimodal correlation feature extraction by disabling the multimodal data correlation feature extraction module (mode-oriented). Second, we evaluate the effectiveness of temporal data feature extraction by disabling the temporal data feature extraction module (time-oriented). Finally, we evaluate the effectiveness of spatial correlation feature extraction by disabling the spatial correlation feature extraction module (node-oriented).

The ablation experiment results are shown in Table 3. We find that disabling the multimodal data correlation feature extraction module leads to a decrease in the F1 score of 3%, disabling the temporal data feature extraction module leads to a decrease in the F1 score of 2.2%, and disabling the spatial correlation feature extraction module leads to a decrease in the F1 score of 1.1%. Disabling the mode direction module and the time direction module leads to a decline in the node expression ability of the node representation vector and affects the subsequent node feature extraction process.

In fact, some modes in the dataset are highly correlated, such as those of temperature and humidity. A graph attention network oriented to the correlation features between multiple modalities helps to accurately capture such correlations for achieving better detection performance. In addition, although the GRU module is used in the model to model long-term dependencies, the graph attention network oriented to the correlation features between multiple times is also very important for the enhancement of performance. One explanation for this phenomenon is that the time-oriented graph attention network can directly model the relationships between nonadjacent timestamps. In this way, some long-term dependencies between timestamps can be more clearly modeled. Compared with the convolution layer, which can only extract information within the size range of the convolution kernel, the graph attention network oriented to the correlation features between the nodes' spatial positions can enable each sensor node to extract the information from all neighbor nodes, so it is more suitable for extracting the features of WSN data flows.

Table 3. Ablation Experiment Results.

| method | Prec | Rec | F1 |
|---|---|---|---|
| original | 93.3% | **87.5%** | **90.3%** |
| -(mode oriented) | 96.0% | 80.0% | 87.3% |
| -(time oriented) | 94.1% | 77.7% | 85.1% |
| -(node oriented) | **96.5%** | 74.5% | 84.0% |

## 6. Conclusions

This paper proposes an anomaly detection method that considers both the temporal and spatial dimensions. This method uses a graph neural network and an attention mechanism to fully extract the temporal and spatial features of a WSN data flow. The experimental results show that this method can effectively detect anomalies in a system in real time when injecting many different types of anomalies, and its detection performance is better than that of the traditional graph convolution network method combined with LSTM. In the next step, the developed method will be optimized under complex conditions involving dataset preprocessing (clearing the irregular points in the dataset in advance), hardware overhead reduction, time complexity reduction, adversarial training and large-scale datasets.

**Acknowledgments**

This research work obtained the subsidization of National Natural Science Foundation of China (Nos. ,62161006, 61811003, 61662018), Guangxi Natural Science Foundation of China (No. 2018GXNSFAA050028), Director Fund project of Key Laboratory of Cognitive Radio and Information Processing of Ministry of Education (Nos. CRKL190102).